# A 10,000-Year Global Stochastic Tropical Cyclone Catalog with Wind-Dependent Track Transitions (WHITS)

Jennifer Nakamura[1] and Upmanu Lall[2,3]

[1]Lamont-Doherty Earth Observatory of Columbia University, Ocean and Climate Physics, Palisades, New York, 10964
https://orcid.org/0000-0003-4113-8519

[2]Arizona State University, School of Complex Adaptive Systems, Tempe, Arizona, 85287
[3]Columbia University, Earth and Environmental Engineering, New York, New York, 10027
https://orcid.org/0000-0003-0529-8128

*Corresponding author*: Jennifer Nakamura, [jennie@ldeo.columbia.edu]

## Abstract

Reliable assessment of tropical cyclone risk is limited by the short and spatially uneven historical record, especially for rare, high-intensity landfalls that dominate insured loss. We present WHITS, the Wind-focused Hurricane Interactive Track Simulator, a non-parametric semi-Markov model that generates a 10,000-year global synthetic catalog of tropical cyclone tracks. WHITS builds new storms by resampling variable-length segments of historical tracks, with segment choices conditioned on local wind speed, location, storm age, and forward motion. This structure preserves both the statistical grounding of the historical record and a form of system memory along realized storm pathways. The resulting catalog reproduces observed track density and the annual probability that hurricane- or typhoon-force winds pass a given location across the major tropical cyclone basins. WHITS is designed to support climate risk analysis, insurance applications, and extreme event research over operational planning time scales by providing a large, low-bias sample of physically plausible storms. This low-bias design is important for applications in which introduced bias cannot simply be corrected after the fact and small errors in storm frequency, track, or intensity can substantially affect loss estimates. Because loss accumulates along the storm path, track geometry matters: storms that loop, stall, or turn sharply can produce damage footprints that differ fundamentally from straight-moving events. By preserving these irregular shapes while greatly expanding the sample of plausible storms, WHITS provides risk analysts, insurers, and coastal planners with information unavailable from observations alone.

## 1 Introduction

Tropical cyclones (TCs) are among the most damaging natural hazards in the world. Just three hurricanes, Harvey, Irma, and Maria in the 2017 North Atlantic season, contributed to global natural-catastrophe losses of approximately 330 billion U.S. dollars overall and 135 billion in insured losses, the highest annual insured total on record at the time (Munich Re 2018). Quantifying the probability of low-frequency, high-severity landfalls is central to insurance pricing, building-code design, and public-sector preparedness, but the historical best-track record is too short and too spatially uneven to support direct empirical estimation of high-return-period events (Hall and Jewson 2007; Emanuel and Jagger 2010). Many coastal regions, particularly outside the North Atlantic, have experienced no high-intensity TC landfall in the available record despite lying within climatologically active corridors, simply because the observational record is short relative to the recurrence interval of the most damaging events.

Stochastic catalogs address this gap by generating long synthetic records that preserve the relevant statistics. Approaches in the literature span fully dynamical models (Emanuel et al. 2006), beta-and-advection schemes with synthetic large-scale flow (Hallegatte 2007), kernel-based track simulators (Hall and Jewson 2007; Yonekura and Hall 2011), and the parametric global model STORM (Bloemendaal et al. 2020). Each represents a different trade-off between physical fidelity, computational cost, and the number of input variables required. For catastrophe-risk applications, where the end use is typically a large Monte Carlo loss estimate rather than an attribution study, what matters most is that the catalog reproduces observed track geometry, intensity, and landfall statistics with low bias, that the sample is large enough for stable tail estimation, and that the underlying method does not impose distributional assumptions that may be violated locally (Emanuel and Jagger 2010).

Bias entering at the catalog-generation stage is particularly costly for risk users. Post-hoc bias corrections to a synthetic catalog typically assume uniformity across regions and storm intensities that is rarely accurate; locally targeted corrections are limited by data sparsity in exactly the regions and intensity bins where corrections matter most; and corrections that ignore dependencies between landfall frequency, intensity, and storm characteristics can produce internally inconsistent loss distributions. Avoiding the introduction of bias upstream is therefore a primary design objective of WHITS: by resampling real historical track segments rather than fitting a parametric distribution, the model inherits the empirical co-variation between track shape, intensity, and landfall geography directly from the observational record, without imposing structural closure assumptions that need to be unwound at the use stage.

HITS (Hurricane Intensity and Track Simulator; Nakamura et al. 2015) was developed in this spirit. It is a non-parametric segment-resampling model inspired by Markov renewal theory (Çinlar 1969). The motivation for going beyond a one-step Markov model came in part from the cluster analysis of Nakamura et al. (2009), which showed that North Atlantic TC tracks fall into a small number of coherent groups that persist over many time steps, with gaps between groups located in specific regions where natural processes cause paths to diverge. Track evolution is therefore not well represented as a process in which each step depends only on the previous one. In HITS, a synthetic track is instead constructed by walking along a randomly chosen historical track segment for a randomly chosen number of steps, then jumping to a kinematically similar neighboring segment, and repeating. By preserving multi-step persistence rather than collapsing it into a one-step transition matrix, HITS avoided the diffusive behavior endemic to spatial Markov-chain track models (Emanuel et al. 2006; Nakamura et al. 2015). It reproduces North Atlantic 6-hour-period counts, mainland landfall counts, and hurricane/typhoon-strength wind statistics across the landfall-box network of Nakamura et al. (2015), with Kolmogorov-Smirnov and Cramer-von Mises two-sample tests passing at the 5% significance level in essentially all landfall boxes. HITS also showed unexpectedly skillful conditional simulation from a single starting position, including for Hurricane Sandy (2012), despite containing no explicit physics.

Two limitations motivated the present work. First, HITS was originally fit to the North Atlantic only. Second, downstream users applying HITS tracks to storm-surge models reported wind-speed discontinuities and small position jumps at segment transitions, which, while statistically harmless for basin-wide diagnostics, produced spurious surge artifacts near coastlines. We therefore developed WHITS, in which W signals both wind-focused and worldwide. The kernel is augmented with a wind-speed term as a stand-alone selector in addition to the existing distance, age, and comparative-vector criteria; the selectivity of the comparative-vector kernel is increased to reduce dissimilar jumps; and a small smoothing window is applied across each transition. WHITS is fit to all six globally significant TC basins using each basin's full available IBTrACS record, taking advantage of the fact that segment resampling does not require a homogeneous observing system in the way parametric model fitting does.

This paper documents the WHITS algorithm, presents a 10,000-yr global synthetic catalog, and validates the catalog against IBTrACS in each basin and against the STORM benchmark (Bloemendaal et al. 2020). The validation focuses on the diagnostics most relevant to catastrophe risk: spatial track density, the climatology of individual seasons, and the annual hurricane/typhoon-force wind-hit probability at each location.

## 2 Data

The historical input is the International Best Track Archive for Climate Stewardship, version 4 (IBTrACS; Knapp et al. 2010), which unifies post-season best-track records from the World Meteorological Organization (WMO) regional specialized meteorological centers and tropical cyclone warning centers. Six basins are retained: the North Atlantic (NA), the Eastern Pacific (EP), the Western Pacific (WP), the North Indian (NI), the South Indian (SI), and the South Pacific (SP). The South Atlantic is excluded because of insufficient activity for stable kernel estimation.

The maximum sustained wind speed reported in IBTrACS (the WMO_WIND field) follows each responsible agency's native averaging convention: 1 min for the National Hurricane Center (NA, EP), 3 min for the India Meteorological Department (NI), and 10 min for the Japan Meteorological Agency, Meteo-France La Reunion, the Australian Bureau of Meteorology, and the Fiji Meteorological Service (WP, SI, SP). To enable consistent kernel comparison and cross-basin display, all wind speeds are converted to a 10-minute maximum sustained wind speed prior to training, applying a factor of 0.88 to the NA and EP 1-min records and 0.93 to the NI 3-min records (Harper et al. 2010); the WP, SI, and SP records require no conversion. All winds reported in the released catalog are on this 10-minute sustained scale. IBTrACS positions and winds are linearly interpolated in time to a uniform 3-hour step prior to training; all references to time steps in the model description below are 3-hourly.

Each basin is fit using its full available best-track record. The North Atlantic and Eastern Pacific records extend furthest, to 1851 and 1876 respectively, reflecting the long ship-based and reanalysis-based extensions performed by HURDAT2 (Landsea and Franklin 2013); the North Indian record extends to 1932; the Western Pacific to 1957; the South Pacific to 1968; and the South Indian to 1973. The records used here run through 2025 in NA, EP, SI, and SP; the WP and NI records used run through 2024, reflecting the IBTrACS release available at the time of catalog generation. Use of each basin's full record, rather than a common modern-era window, is a deliberate design choice: WHITS resamples segments rather than fitting parametric distributions, so each additional year of record enriches the segment library available for conditional sampling without requiring observational homogeneity (Halevy et al. 2009; Nakamura et al. 2015). Earlier tracks are nonetheless known to be less complete in the deep tropics and for weaker systems (Landsea et al. 2010; Schreck et al. 2014), and we therefore restrict the genesis-position sampling distribution and the season-length sampling distribution to the modern observing-era subsets of each basin's record. The cutoff years used are 1944 for NA, WP, and SI (coincident with the start of routine aircraft reconnaissance in the North Atlantic; Hall and Jewson 2007), 1945 for EP, 1951 for NI, and 1968 for SP, in each case reflecting the onset of routine basin-wide recordkeeping at the responsible warning center. The much longer historical segment libraries are still drawn upon for the kernel-based transitions during simulation, ensuring that the catalog inherits the full range of observed track shapes while genesis statistics are anchored to the modern observing era.

## 3 Method

### 3.1 Framework

WHITS treats each historical track as a realization of a hidden environmental-flow state and constructs synthetic tracks by walking along historical segments and probabilistically transitioning to kinematically similar neighboring segments. Formally, this is a non-homogeneous hidden Markov renewal model (Çinlar 1969; Nakamura et al. 2015): the time spent in any one latent state is random, transitions between states occur at randomly selected candidate locations, and the conditional probability of transitioning to a particular neighboring segment depends on local covariates. This structure preserves multi-step persistence and avoids the diffusive behavior of one-step Markov-chain track models (Emanuel et al. 2006; Nakamura et al. 2015).

A simulation begins by sampling a genesis location from the empirical distribution of historical genesis points within the basin's modern observing-era subset and a season length from the corresponding empirical distribution. The track then steps forward in 3-hour increments along its current host segment. At each step the algorithm decides whether to remain on the current segment or transition to a neighbor. If a transition is selected, the destination segment is drawn from the set of candidate segments inside a 2.5° radius of the current position, weighted by the kernel described in section 3.2. The track terminates when its sampled lifetime is reached.

### 3.2 Wind-conditioned kernel

Following the HITS framework (Nakamura et al. 2015, their eqs. 5a–5e), a synthetic track is advanced by repeatedly transitioning from the current host segment to a kinematically similar neighboring segment at a candidate transition location $\boldsymbol{x}^*$. Let $C(\boldsymbol{x}^*)$ denote a candidate historical track passing through $\boldsymbol{x}^*$, $\tau[C(\boldsymbol{x}^*)]$ the random time, in 3-hour steps, spent along that track once it is selected, and $\boldsymbol{\theta}(\boldsymbol{x}^*)$ the vector of local covariates governing the transition at $\boldsymbol{x}^*$. The conditional density for selecting a candidate neighbor track is built as a product of bisquare kernels acting on standardized covariates that compare the current (arriving) track $i$ with a candidate neighbor track $j$. We first define the kernel and the three covariates, then combine them into the conditional density. Each covariate is passed through a generalized bisquare kernel

$$K(u;\alpha) = (1-u^2)^{\alpha}, \qquad |u| \leq 1 \tag{1}$$

where $u$ is a standardized covariate, $\alpha \geq 1$ is a shape parameter that controls how sharply the kernel penalizes large covariate values, and $K(u; \alpha) = 0$ for $|u| > 1$. In HITS (Nakamura et al. 2015), $\alpha = 2$ was used for all three kernel terms, recovering the standard bisquare. The three standardized covariates are defined between the arriving track $i$ and the candidate neighbor track $j$ at the transition location, following Nakamura et al. (2015, their eqs. 5c–5e). The first is a spatial distance,

$$u_{1,ij} \propto D_{ij} \cdot 2.5\left(\frac{\pi}{180}\right), \qquad D_{ij} \leq 2.5^{\circ} \tag{2}$$

where $D_{ij}$ is the great-circle distance in degrees, in spherical geometry, between the current position on track $i$ and the candidate point on track $j$; only candidates within the 2.5° search radius are retained [Nakamura et al. (2015), their eq. 5c]. The second is a comparative wind-vector difference,

$$u_{2,ij} \propto \frac{V}{\max(V)} \tag{3}$$

where $V = \left[\left(V_{x,j} - V_{x,i}\right)^2 + \left(V_{y,j} - V_{y,i}\right)^2\right]^{1/2}$ is the magnitude of the difference between the local wind vectors of the two tracks. For each track the components $V_x$ and $V_y$ equal the maximum sustained wind speed multiplied by the cosine and sine, respectively, of the local forward-motion (heading) angle; subscripts $i$ and $j$ denote the arriving and candidate tracks, and $\max(V)$ is the maximum of $V$ over all candidate tracks in the basin [Nakamura et al. (2015), their eq. 5d]. The third is a relative age,

$$u_{3,ij} \propto \frac{T}{\max(T)} \tag{4}$$

where $T = T_j - T_i$ is the difference between the ages of the candidate and arriving tracks, $T_i$ and $T_j$ are those ages measured in 3-hour steps from genesis, and max($T$) is the maximum of $T$ over all candidate tracks in the basin [adapted from Nakamura et al. (2015), their eq. 5e, with the time-step interval changed from 6 h to 3 h to match the 3-hourly interpolation used here]. Combining the kernel (1) with these three covariates, the HITS conditional density is

$$f\{C(\mathbf{x}^*),\ \tau[C(\mathbf{x}^*)] \mid \boldsymbol{\theta}(\mathbf{x}^*)\} \propto K(u_{1,ij})\, K(u_{2,ij})\, K(u_{3,ij}) \quad (5)$$

WHITS modifies (5) in two ways. First, although the comparative wind-vector covariate $u_2$ nominally couples wind speed to forward direction, in practice candidate neighbors that match in direction can still differ substantially in speed magnitude, producing the discontinuous wind-speed steps observed in HITS at transitions. We therefore introduce a fourth covariate that depends on wind-speed magnitude alone,

$$u_{4,ij} \propto \frac{|w_j - w_i|}{\max(|w_j - w_i|)} \quad (6)$$

where $w_i$ and $w_j$ are the IBTrACS 10-minute maximum sustained wind speeds on the arriving and candidate tracks at the matching time step, and the normalization is the basin maximum of the absolute speed difference. Second, the shape parameter $\alpha$ in (1) is increased above 2 for the comparative-vector and wind-magnitude terms, sharpening the kernel and penalizing candidates with mismatched motion vectors or speed more strongly. The shape parameter for the distance and age terms is held at $\alpha = 2$, matching HITS. The combined WHITS conditional density is

$$f\{C(\mathbf{x}^*),\ \tau[C(\mathbf{x}^*)] \mid \boldsymbol{\theta}(\mathbf{x}^*)\} \propto K(u_{1,ij})\, K(u_{2,ij})\, K(u_{3,ij})\, K(u_{4,ij}) \quad (7)$$

with $K(\cdot)$ the kernel of (1) applied to each covariate, with $\alpha = 2$ for the distance and age terms (matching HITS) and $\alpha > 2$ for the comparative-vector and wind-magnitude terms. Adding wind speed as a separate, magnitude-only term gives speed its own selection pressure independent of direction; this primarily reduces wind-speed mismatch at transitions while leaving the spatial track-density statistics that HITS reproduces well largely unchanged. The value of $\alpha$ used for the comparative-vector and wind-magnitude terms is selected by iterative regeneration of the 10,000-year catalog and direct visual comparison of the resulting simulated tracks and aggregate fields against IBTrACS across three diagnostics: track smoothness and the absence of artifacts at segment transitions in representative simulated seasons (cf. Fig. 1), the per-year track density (cf.

Fig. 2), and the annual hurricane/typhoon-force wind-hit probability (cf. Fig. 3). A single value of $\alpha$ is applied uniformly to all six basins, rather than tuned per basin; this is a deliberate choice intended to keep the model parsimonious and to test whether the kernel structure (as opposed to per-basin parameter tuning) is what drives the agreement with observations. Because the global parameter setting is selected on the same diagnostics presented as validation in section 4, the close agreement between the simulated and observed fields in Figs. 1, 2, and 3 is partly by construction. The fact that a single setting reproduces observed track density and hurricane/typhoon-force wind-hit probability across six basins with very different climatological corridors and intensity distributions, however, suggests that the kernel structure rather than the parameter values is doing most of the work, and serves as a partial out-of-sample check on the underlying segment-resampling methodology. The benchmark comparison with STORM in section 5 provides additional context.

### 3.3 Transition smoothing

Even with the sharpened kernel, instantaneous segment transitions can leave small position offsets and wind-speed steps at the join. Both are statistically negligible at the basin scale but matter for downstream surge modeling, where a 5 m $s^{-1}$ step in 10-m wind translates to a discontinuous storm-surge forcing. WHITS therefore handles transitions in two stages. First, the destination segment is translated so that its first point coincides with the current track position, eliminating the position jump. Second, the latitude, longitude, and wind-speed series are linearly smoothed across a short window centered on each join, removing the step in wind speed and any residual position offset. The smoothing window is short enough that segment-level statistics are preserved (verified by recomputing kernel-fit diagnostics on smoothed versus raw catalogs) and long enough to remove the discontinuities flagged by surge users. Smoothing is applied only in this five-point transition window and is never applied to the body of a segment, so the natural variability of historical track shape (recurvatures, sharp turns, hairpins, loops, and other unique geometries) is preserved unaltered in the simulated catalog.

Transitions are not permitted at the final point of a source segment, so that each join has room for the destination segment to be appended. The catalog generator reserves a small basin-specific number of time steps for this purpose, scaling with each basin's mean track length.

### 3.4 Catalog generation

All transition probabilities from every track point to every candidate point are precomputed once and stored in a lookup table, after which catalog generation is fast: 10,000 simulated years per basin run in under a day on a multi-core workstation. Retraining the precomputed probabilities to incorporate additional best-track data takes on the order of days of compute. Because the kernel-based jump step is stochastic, individual catalog runs differ in detail, but bulk statistics over a 10,000-yr catalog are stable. The complete code, configuration files, and the 10,000-yr catalog used in this paper are available at the project repository.

## 4 Validation

### 4.1 Approach to cross-basin comparison

The six basins are observed for different lengths of time, ranging from 53 yr (SI) to 175 yr (NA). To make the fields directly comparable across basins, all observed and simulated fields are normalized to a per-year rate by dividing by $N_b$, the basin-specific number of years used to compute the field. For consistency with the modern observing-era genesis-sampling subsets defined in section 2, the observed reference fields shown in Figs. 2 and 3 are computed over the modern observing-era window of each basin (NA and WP from 1944, EP from 1945, NI from 1951, SP from 1968, and SI from 1973), all extending through 2024 or 2025 as available. Simulated fields are computed by drawing 100 random samples of $N_b$ years from the 10,000-yr catalog, computing the per-year rate within each sample, and retaining the cell-wise sample median as a robust central estimate (Emanuel and Jagger 2010). Because WHITS resamples segments pooled across the full training period, the simulated catalog itself is not bound to any particular observational window, and the same 10,000-yr run serves all six basins. Exact cell-by-cell agreement between the observed and simulated fields is neither expected nor a meaningful target. The observed field for any basin is a single realization drawn from $N_b$ years of nature, while the simulated field is the median across many independent $N_b$-year draws from the catalog. Even a simulator that perfectly captured the underlying climatology would differ from the observed field at the cell level by an amount set by the sampling variability of $N_b$ years. What matters is whether the simulated and observed fields agree in their broad spatial patterns and overall rates, not whether they match cell-by-cell.

### 4.2 Individual simulated seasons

Fig. 1 compares five seasons of historical IBTrACS tracks (2020-2024) against the first five simulated seasons from the WHITS catalog, with the basins grouped by IBTrACS basin definitions onto three panel rows to minimize blank ocean. Each track is colored by its 10-minute maximum sustained wind speed and originates at a marked genesis point. The two samples are not expected to match each other in detail: both are five-year subsets, the historical window happens to be an above-average period in the North Atlantic, and the first five simulated seasons are an arbitrary slice of the catalog. The comparison is qualitative, intended to demonstrate that the simulated tracks reproduce the features expected of TC climatology: the parabolic recurvature characteristic of the North Atlantic main development region (Elsner and Kara 1999), the westward and northwestward propagation of Western Pacific tracks, the relatively short and meridionally constrained tracks of the North Indian, and the southward-then-eastward arcs of the Southern Hemisphere basins. The simulated tracks are visually similar in spatial coverage and intensity distribution to the historical comparison sample. Because the segment-shift mechanism in section 3.3 can translate a candidate segment to begin at a position not previously occupied by that segment, simulated tracks can occupy combinations of position and motion that have not been observed in the historical record while still aggregating to the correct basin-scale statistics. This addresses a limitation noted in the original HITS, in which simulated tracks could only retrace historical segments exactly. No track exhibits unphysical reversals, and the wind-speed evolution along each track is smooth at the segment-transition scale.

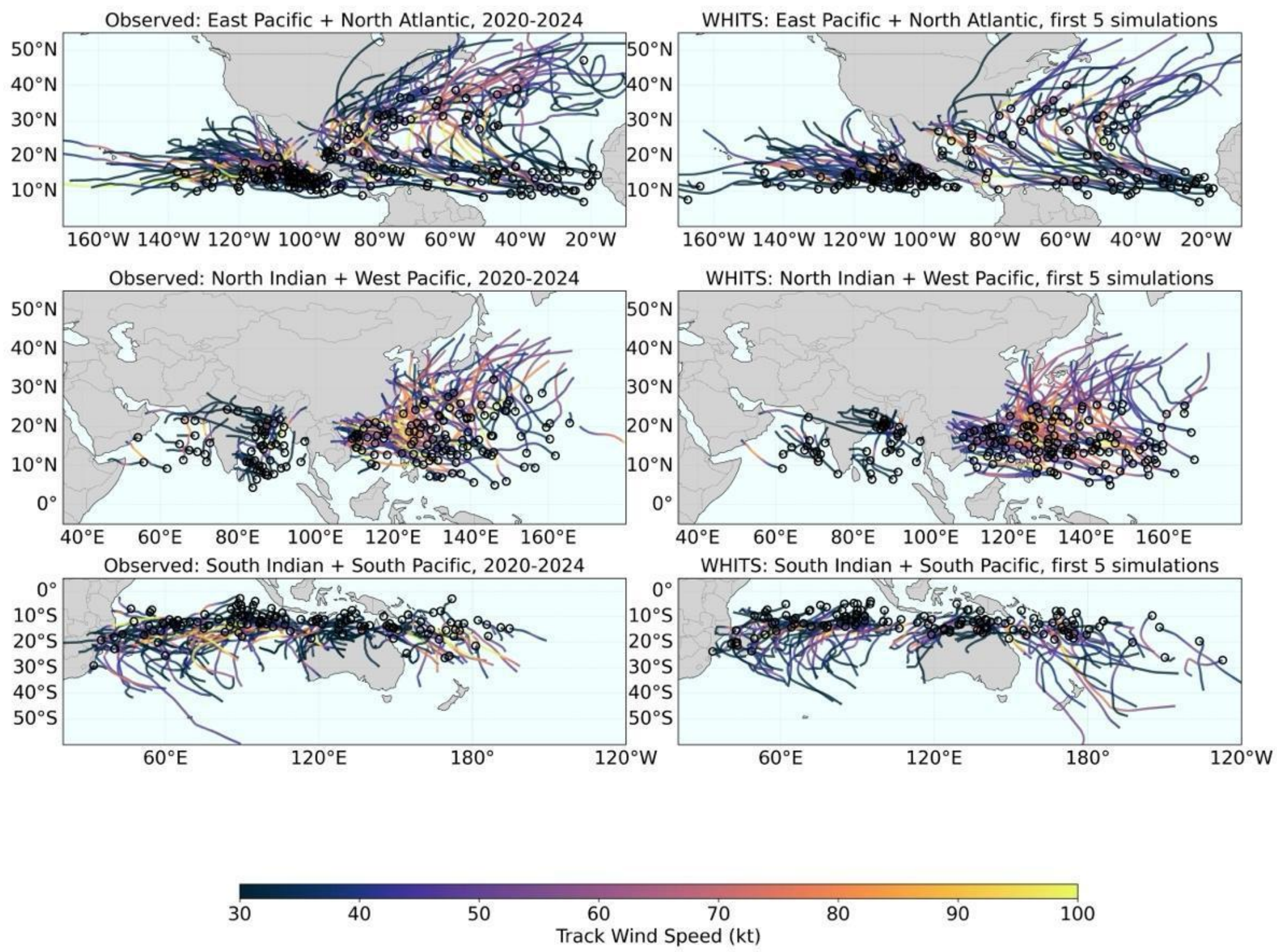

**Fig. 1** Comparison of five seasons of historical IBTrACS tracks (left columns, 2020-2024) and the first five simulated seasons from the 10,000-yr WHITS catalog (right columns), with each track beginning at an open circle (genesis point) and colored by 10-minute maximum sustained wind speed. To minimize blank ocean, basins are grouped onto three panel rows using the IBTrACS basin definitions: (top) Eastern Pacific and North Atlantic combined; (middle) North Indian and Western Pacific combined; (bottom) South Indian and South Pacific combined. Both the historical (2020-2024) and simulated samples shown are five-year subsets and are not expected to match each other in detail; 2020-2024 was an above-average period in the North Atlantic in particular, and the first five simulated seasons are an arbitrary starting slice of the catalog. The figure illustrates qualitative track-shape realism (basin-appropriate corridors, recurvature, occasional loops and hairpins) and the absence of unphysical artifacts at segment transitions following the wind-conditioned kernel and transition smoothing introduced in WHITS.

The transition smoothing of section 3.3 is applied only locally, in a short window centered on each segment-to-segment join, and never across the body of a segment. The interior of every segment is the historical track itself, and the natural variability of TC shape is therefore preserved unaltered in the simulated catalog. This matters because real tracks routinely exhibit shapes that any heavily smoothed track generator would erase. Examples in the historical record include sharp 90° turns (e.g., North Atlantic: Charley 2004, Wilma 2005, Babe 1977, Elena 1985, Michael 2018, Franklin 2023; Western Pacific: In-fa 2021, Lionrock 2016, Jongdari 2018; Central Pacific: Walaka 2018), hairpin reversals (North Atlantic: Ingrid 2013, Joaquin 2015, Alice 1954-55, Idalia 2023; Western Pacific: Noru 2017), loops (North Atlantic: Jeanne 2004, Betsy 1965, Danielle 2022, Don 2023, Nigel 2023, Katia 2023, Margot 2023, Rafael 2024; Western Pacific: Wayne 1986, Opal 1962, Vera 1959), and one-of-a-kind geometries (all North Atlantic) such as the pretzel of Tropical Storm Gert 2023, the bow of Hurricane Juan 1985, the figure-eight of Hurricane Leslie 2018, and the leaf of Tropical Storm Karl 2022. Because WHITS resamples segments rather than parameterizing track geometry, all such shapes appear in the simulated catalog with frequencies consistent with their occurrence in IBTrACS, providing an unbiased sample of natural TC behavior.

### 4.3 Track density

Fig. 2 compares the observed and simulated track density on a 2° × 2° grid, expressed as track-points per year per grid cell. The fields are displayed on a logarithmic color scale because track-point rates span several orders of magnitude across each basin (from order 1 in the densest corridors to $10^{-3}$ to $10^{-4}$ in the far field); a linear scale would render everything outside the corridor maxima as apparent zero. Track density is the primary diagnostic for kernel-based track models: realistic track density is a near-sufficient condition for realistic landfall statistics (Hall and Jewson 2007; Nakamura et al. 2015), and its spatial pattern encodes the steering-flow climatology that any acceptable model must reproduce.

The observed and simulated fields are visually indistinguishable in all six basins. The North Atlantic shows the broad arc around the climatological subtropical high, with a dense corridor through the Caribbean and the Bahamas and a recurvature track east of the United States; the Eastern Pacific shows the long zonal corridor west of Mexico, with the highest densities right off the Mexican coast; the Western Pacific shows the broad maximum east of the Philippines and the secondary corridor through the South China Sea; the North Indian shows the bimodal Bay-of-Bengal and Arabian-Sea structure; the South Indian shows the long zonal corridor extending east from Madagascar across the basin; and the South Pacific shows the Coral Sea maximum and the eastward extension toward Fiji. In every basin, the dominant corridors and density maxima of the simulated field align with those of IBTrACS in both location and orientation. The simulated field is also smoother than the observed field, particularly in the far-field regions where isolated, noisy cells in the observed record are filled in by the much larger effective sample size of the synthetic catalog (10,000 yr versus 53-82 yr per basin).

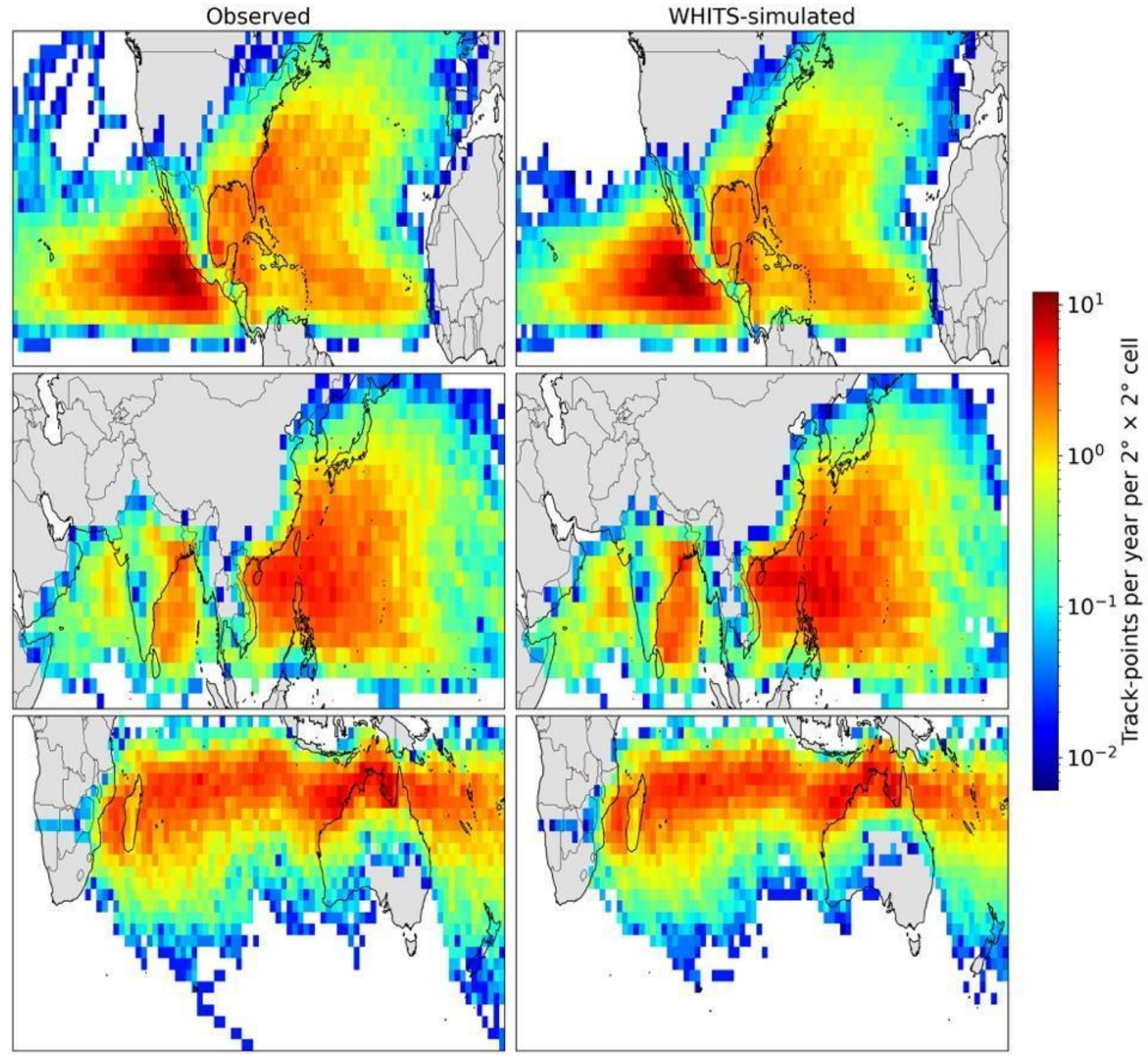

**Fig. 2** Observed (left columns) and WHITS-simulated (right columns) tropical cyclone track density on a 2° × 2° grid, expressed as track-points per year per grid cell. To minimize blank ocean, basins are grouped onto three panel rows using the IBTrACS basin definitions: (top) Eastern Pacific and North Atlantic combined; (middle) North Indian and Western Pacific combined; (bottom) South Indian and South Pacific combined. Observed fields are computed from IBTrACS over each basin's modern observing-era subset (NA 1944-2025, $N_b$ = 82; EP 1945-2025, $N_b$ = 81; WP 1957-2024, $N_b$ = 68; NI 1951-2024, $N_b$ = 74; SP 1968-2025, $N_b$ = 58; SI 1973-2025, $N_b$ = 53) and divided by $N_b$. Because WHITS resamples segments pooled across the full training period rather than reproducing a specific observational window, the same 10,000-yr catalog serves all basins; the simulated field is obtained by drawing 100 random samples of $N_b$ years per basin, computing the per-year rate in each draw, and retaining the cell-wise median. The color scale is logarithmic. Track-density values span several orders of magnitude across each basin: dense climatological corridors carry on the order of 1 track-point per year per cell, while far-field regions carry rates as low as $10^{-3}$ to $10^{-4}$ per year per cell. A linear scale would saturate the corridor maxima and reduce the broader field to apparent zero, obscuring the low-but-nonzero density that constitutes the climatological reach of each basin and that is relevant for catastrophe-risk applications. The logarithmic scale preserves visible structure across the full dynamic range.

### 4.4 Annual hurricane/typhoon-force wind-hit probability

Track density is not, by itself, a complete validation: a model could reproduce the spatial distribution of track points while badly mis-specifying intensity along those tracks. To probe intensity, we compute for each grid cell the empirical probability that the cell experiences hurricane/typhoon-force conditions at least once during a year. A grid cell is counted as a hit in a given year when the sum of wind speeds at all track points falling inside the cell reaches or exceeds 64 kt (33 m $s^{-1}$, the hurricane/typhoon-force threshold):

$$P_{64}(\text{cell}) = \frac{n_{64}}{N} \tag{8}$$

Here $n_{64}$ is the number of years in which the summed wind speed in the cell reaches or exceeds 64 kt, as defined above, and $N$ is the total number of years. This wind-hit probability has a direct risk-management interpretation: a value of 0.4 means a four-in-ten chance that the cell experiences hurricane/typhoon-force conditions in any given season. This metric sums wind speeds rather than taking a maximum, so it weights cells by both storm intensity and time spent in the cell. For risk applications where annual cumulative wind exposure matters, this is closer to a relevant quantity than a simple maximum-wind exceedance. Fig. 3 compares observed and simulated $P_{64}$ on the 2° × 2° grid; the simulated field is the cell-wise median of $P_{64}$ computed independently on 100 random $N_b$-yr draws from the catalog. As in Fig. 2, the fields are displayed on a logarithmic color scale; $P_{64}$ spans roughly two orders of magnitude across each basin, and a linear scale would mask low-probability hurricane/typhoon-force wind-hit cells that nonetheless contribute meaningfully to long-term catastrophe risk.

The agreement is close in every basin. The Western Pacific shows the highest probabilities of the six basins, with a broad maximum east of Luzon extending through the East China Sea and the Philippine Sea; the Eastern Pacific peaks in the corridor west of southern Mexico and Baja California; the North Atlantic shows a spatially extended pattern with the strongest values across the western Caribbean and the western Atlantic; the Southern Hemisphere basins show their maxima in a long zonal band across the central South Indian and across the Coral Sea region; and the North Indian shows comparatively narrow peaks at the head of the Bay of Bengal and over the Arabian Sea. In every basin the spatial pattern of the simulated $P_{64}$ matches the observed pattern, and the magnitudes agree across the dynamic range of the field. Because the WHITS catalog contains 10,000 yr rather than the 53-82 yr available from each basin's modern observing-era IBTrACS subset, the simulated field is also smoother than the observed field, which is the desired behavior for a catalog intended to support stable estimation of long return periods.

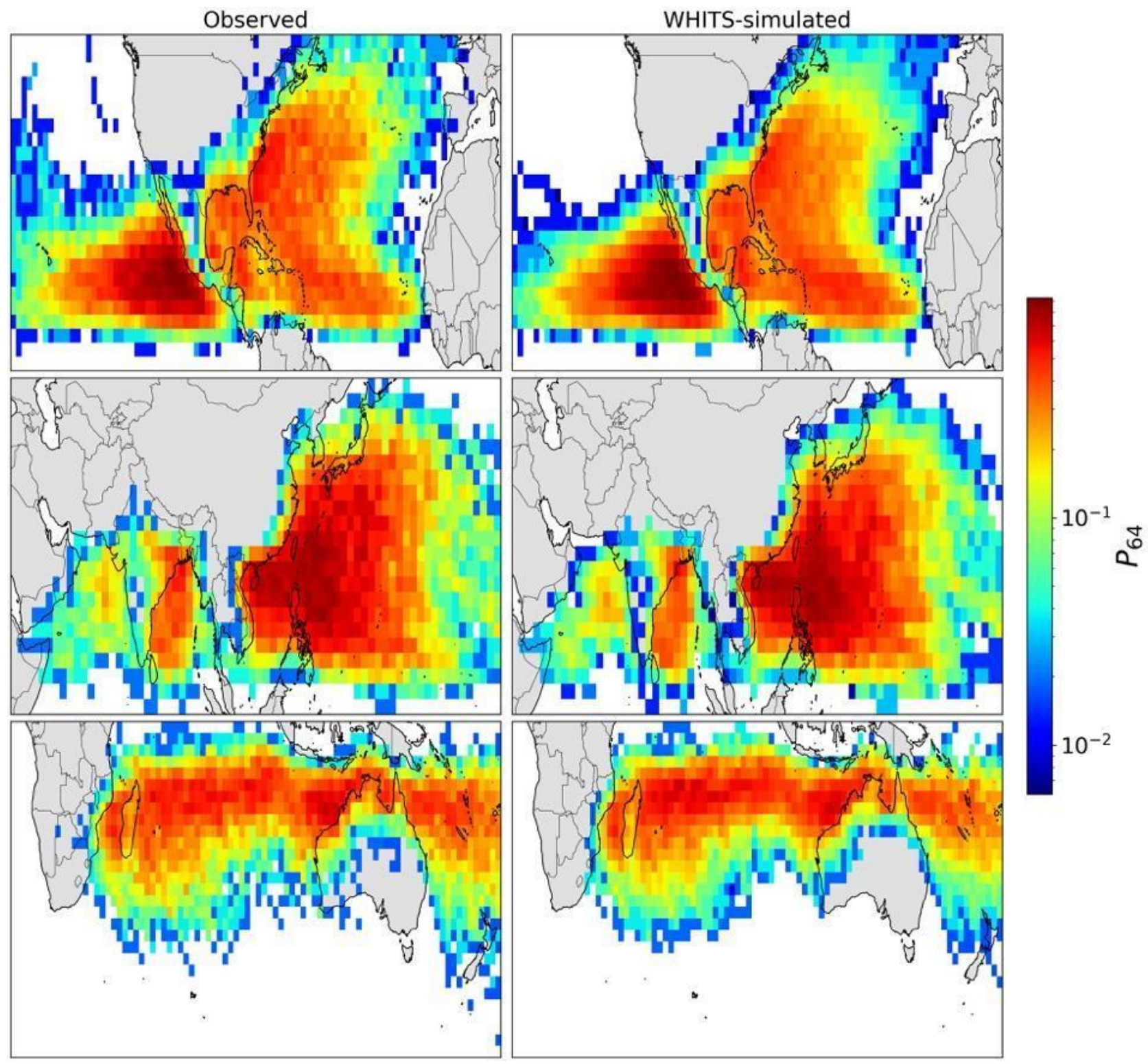

**Fig. 3.** Observed (left columns) and WHITS-simulated (right columns) annual hurricane/typhoon-force wind-hit probability, $P_{64}$, on a 2° × 2° grid. For each year, a grid cell is counted as a hit when the sum of wind speeds at all track points falling inside the cell reaches or exceeds 64 kt (33 m $s^{-1}$). Basins are grouped as in Fig. 2: (top) Eastern Pacific and North Atlantic; (middle) North Indian and Western Pacific; (bottom) South Indian and South Pacific. Observed fields use each basin's modern observing-era IBTrACS subset (years as in Fig. 2). The simulated field is the cell-wise median of $P_{64}$ computed from 100 random $N_b$-yr draws from the WHITS catalog. Values are annual probabilities, with 0.4 corresponding to a four-in-ten chance per season. The color scale is logarithmic. $P_{64}$ spans roughly two orders of magnitude across each basin, from values approaching 0.6 in the densest corridors to values below 0.01 in the broader hurricane/typhoon-affected region. A linear scale would saturate the corridor maxima and reduce the broader field to apparent zero, masking low-probability hurricane/typhoon-force wind-hit cells that nonetheless contribute meaningfully to long-term catastrophe risk.

## 5 Benchmark comparison with STORM

STORM (Bloemendaal et al. 2020) is included as a benchmark because it is the most directly comparable global synthetic TC catalog in the published literature, is publicly available, and is widely used in the catastrophe-risk community. STORM and WHITS share the same IBTrACS source and the same 10,000-yr catalog length, which allows the two catalogs to be evaluated against the same observational reference using the same diagnostics introduced in section 4.

Figs. 4-6 reproduce the same three diagnostics for the public STORM catalog: representative simulated seasons (Fig. 4), observed-versus-simulated track density (Fig. 5), and observed-versus-simulated annual hurricane/typhoon-force wind-hit probability (Fig. 6). Per-year normalization follows the same convention as Figs. 2 and 3, with both observed and simulated fields constructed against each basin's modern observing-era IBTrACS window. Figs. 1-3 and 4-6 can therefore be compared side by side across the same diagnostics and basins.

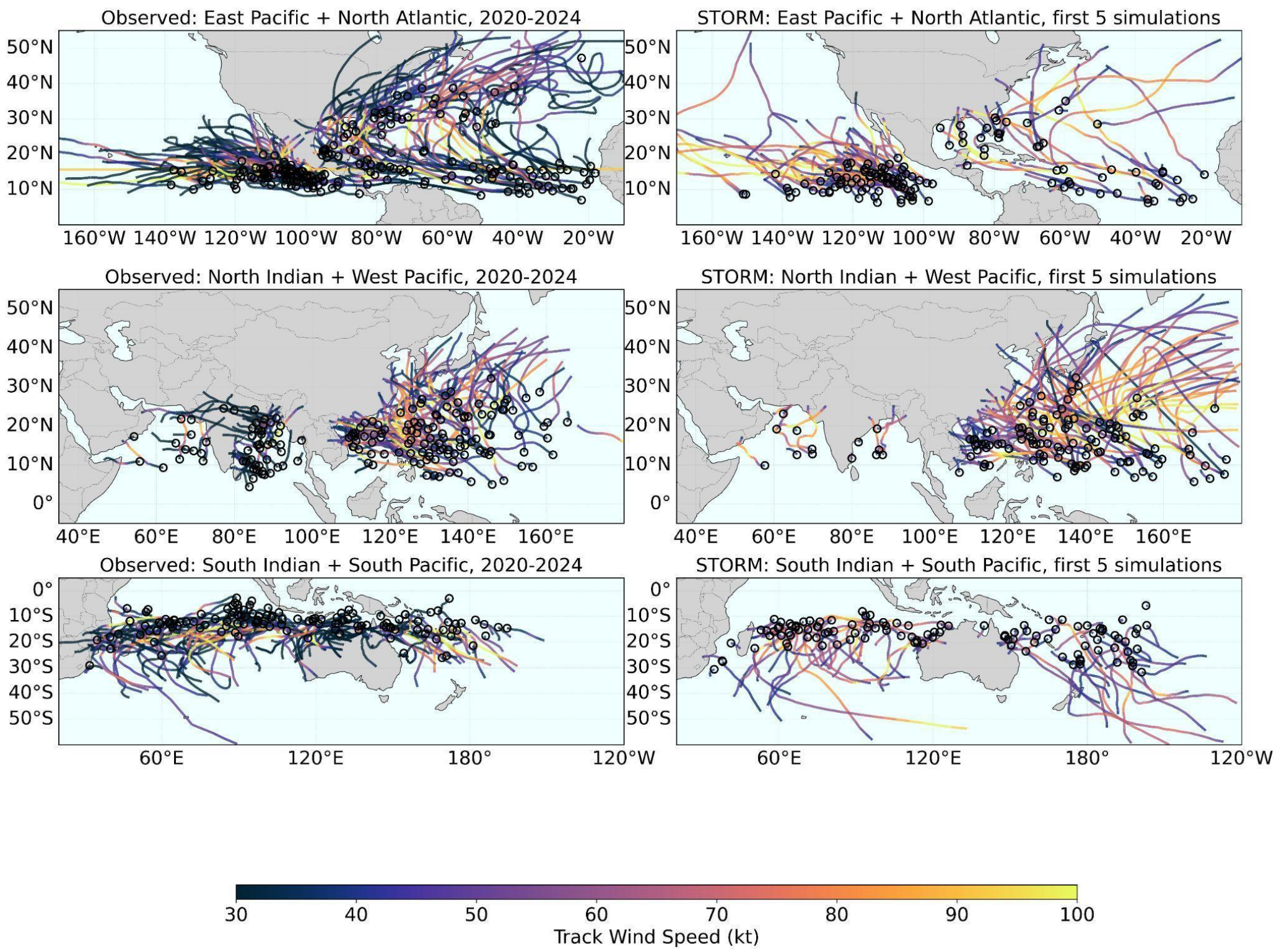

**Fig. 4** As Fig. 1, but with the simulated tracks (right columns) drawn from the first five years of the public STORM 10,000-yr catalog (Bloemendaal et al. 2020) instead of WHITS, shown for benchmark comparison. The same three-panel grouping by IBTrACS basin definitions is used: (top) Eastern Pacific and North Atlantic; (middle) North Indian and Western Pacific; (bottom) South Indian and South Pacific. STORM is a parametric model fit to IBTrACS over 1980-2018.

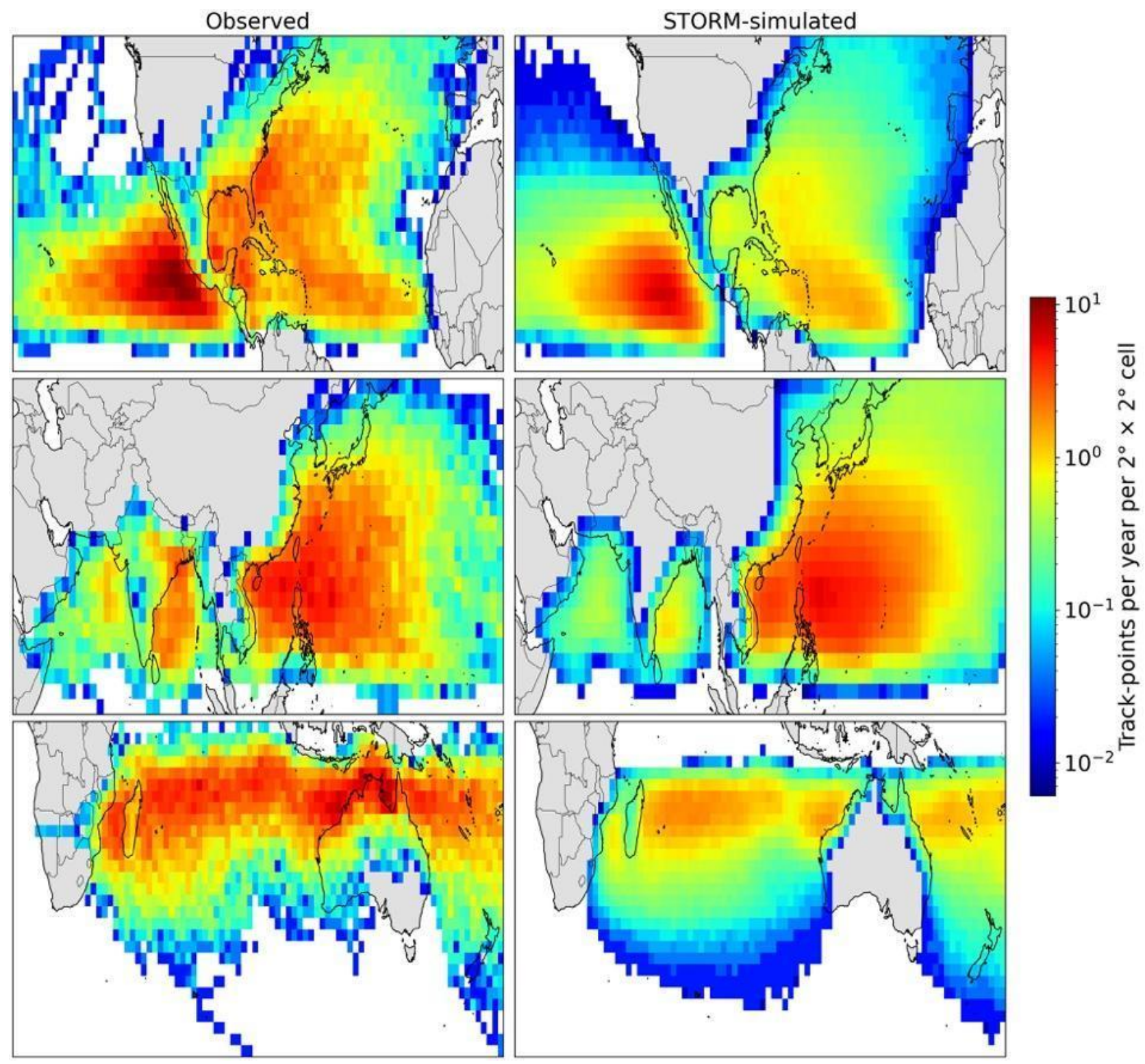

**Fig. 5** As Fig. 2, but for the STORM benchmark, using the same three-panel basin grouping. Observed reference fields and $N_b$ values are identical to those of Fig. 2; simulated fields are computed from the public STORM catalog by drawing 100 random $N_b$-yr samples per basin, computing the per-year rate within each sample, and retaining the cell-wise median.

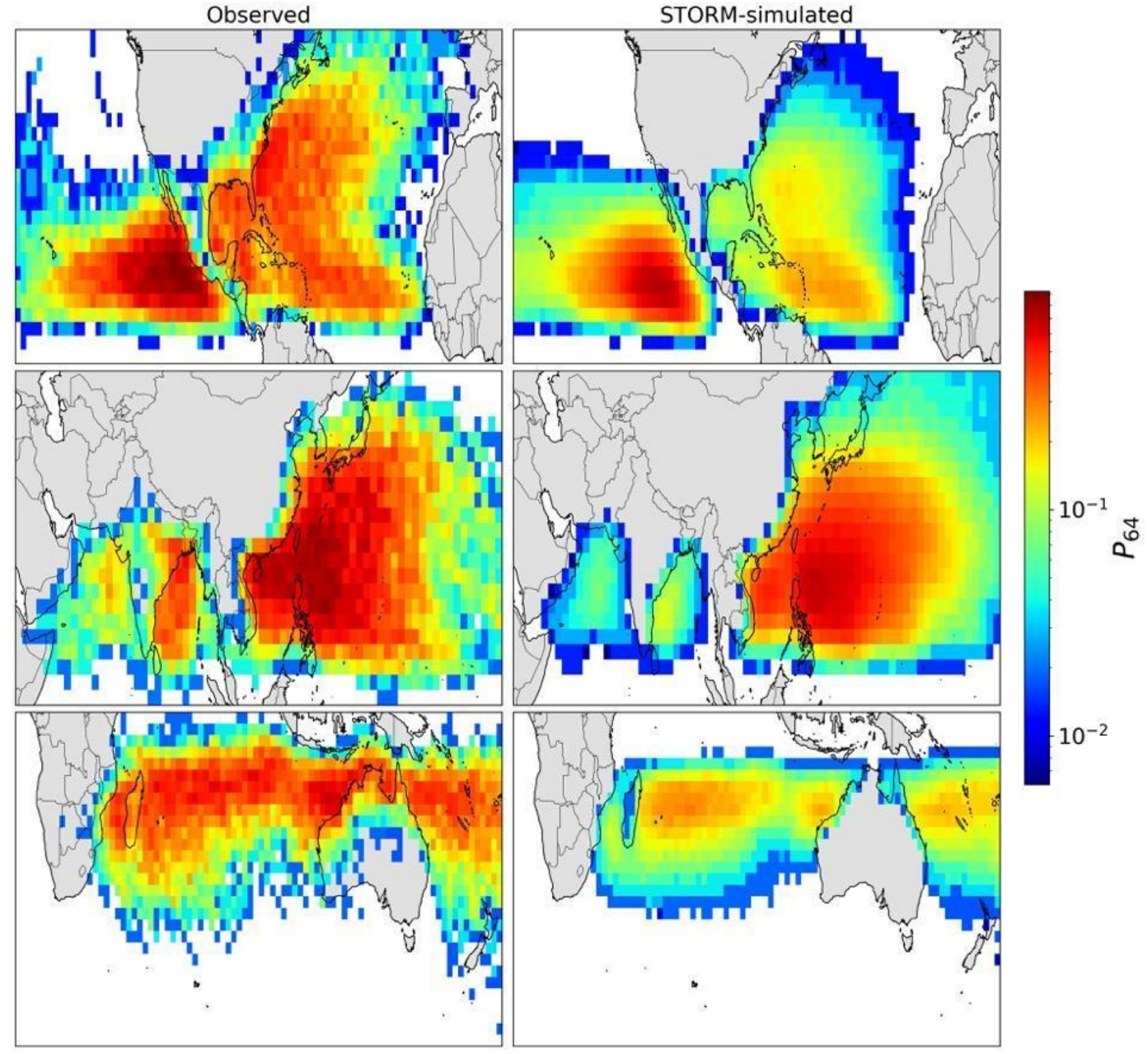

**Fig. 6** As Fig. 3, but for the STORM benchmark, using the same three-panel basin grouping. The observed reference field is identical to that of Fig. 3; the simulated field is the cell-wise median of $P_{64}$ computed independently on 100 random $N_b$-yr draws from the public STORM catalog.

## 6 Applications and discussion

The 10,000-yr WHITS catalog is intended primarily for catastrophe-risk applications. The two classes of use case we anticipate are (i) loss estimation, in which the catalog serves as the hazard input to a vulnerability and exposure model and the resulting loss distribution is characterized at long return periods; and (ii) coastal-engineering and design-wind estimation, in which the catalog supports return-period estimates of peak gust or sustained wind at specific points. The smoothed transitions introduced in WHITS make output (ii) and the surge-modeling extension of (i) substantially more reliable than was possible with HITS.

The framework is targeted at present-climate risk assessment and is designed to support climate-risk analysis, insurance applications, and extreme-event research on operational planning time scales of approximately five years, over which the underlying climatology is expected to remain a reasonable approximation to the present-day environment. Updates beyond that horizon are straightforward in principle: retraining on an updated IBTrACS release brings the segment library and genesis distribution into alignment with the current observing record, with the most recent retraining (incorporating IBTrACS through the 2025 season) completed in approximately two days of compute.

## 7 Conclusions

WHITS extends the non-parametric segment-resampling track simulator HITS (Nakamura et al. 2015) to all six globally significant TC basins, adds a wind-speed term to the segment-transition kernel, sharpens the comparative-vector selectivity, and applies a short smoothing window across transitions to remove position and wind-speed discontinuities. fit to each basin's full IBTrACS record, extending in the North Atlantic to 1851, and with genesis sampling restricted to the modern observing-era subset of each basin, the model produces a 10,000-yr global synthetic catalog that reproduces observed track density and the annual hurricane/typhoon-force wind-hit probability in every basin and that compares favorably with the parametric STORM benchmark on the diagnostics relevant to catastrophe-risk applications. The catalog is provided for use in loss estimation and design-wind work.

## Statements & Declarations

### Acknowledgments

This research received no external funding. The authors thank Columbia Technology Ventures for managing the licensing arrangements. We gratefully acknowledge insightful discussions with industry experts that contributed to model refinement.

### Data Availability Statement

The 10,000-yr WHITS catalog, together with the training code and documentation, is available under a license agreement administered by Columbia Technology Ventures. A free click-through license for academic, non-commercial research is available through the WHITS Downloadable Package: https://inventions.techventures.columbia.edu/technologies/whits-downloadable-package--CU25254. After completing the academic non-commercial license, users should email the corresponding author, Jennifer Nakamura (jennie@ldeo.columbia.edu), with a brief 2–3 sentence description of their planned use. This helps track users, understand emerging applications, and provide relevant updates as needed. For commercial licensing inquiries, please contact Dovina Qu at Columbia Technology Ventures (techtransfer@columbia.edu) and copy Jennifer Nakamura (jennie@ldeo.columbia.edu). WHITS is © 2026 The Trustees of Columbia University in the City of New York. Additional information is available at http://rainbow.ldeo.columbia.edu/~jennie/WHITS.

The IBTrACS dataset (Knapp et al. 2010) is publicly available from NOAA's National Centers for Environmental Information at https://www.ncei.noaa.gov/products/international-best-track-archive. The public STORM catalog (Bloemendaal et al. 2020) used for benchmark comparison is available at https://doi.org/10.4121/uuid:82c1dc0d-5485-43d8-901a-ce7f26cda35d.

**Competing Interests**

The WHITS dataset is licensed through Columbia Technology Ventures on behalf of Columbia University. Academic, non-commercial use is free, but future revenue from commercial use or commercial licensing could help support Jennifer Nakamura’s soft-money research program.

Upmanu Lall declares no competing interests.

**Author Contributions**

All authors contributed to the study conception and design. Material preparation, data collection and analysis were performed by Jennifer Nakamura. The first draft of the manuscript was written by Jennifer Nakamura and all authors commented on previous versions of the manuscript. All authors read and approved the final manuscript.